\documentclass[preprint]{elsarticle}

\usepackage{lineno,hyperref}
\modulolinenumbers[5]
\usepackage{amssymb}
\usepackage{url}
\usepackage{xcolor}
\usepackage{graphicx}
\usepackage[cmex10]{amsmath}
\usepackage{multirow}
\usepackage{geometry}
\journal{Journal of Image and Vision Computing}

\begin{document}

\begin{frontmatter}

\title{\huge\textbf{Simplified~Active~Calibration}}

\author{Mehdi Faraji}
\ead{faraji@ualberta.ca}
\author{Anup Basu \corref{mycorrespondingauthor}}
\cortext[mycorrespondingauthor]{Corresponding author}
\address{Department of Computing Science, University of Alberta, Canada}
\ead{basu@ualberta.ca}

%
%

\begin{abstract}
We present a new mathematical formulation to estimate the intrinsic parameters of a camera in active or robotic platforms. We show that the focal lengths can be estimated using only one point correspondence that relates images taken before and after a degenerate rotation of the camera. The estimated focal lengths are then treated as known parameters to obtain a linear set of equations to calculate the principal point. Assuming that the principal point is close to the image center, the accuracy of the linear equations are increased by integrating the image center into the formulation. We extensively evaluate the formulations on a simulated camera, 3D scenes and real-world images. Our error analysis over simulated and real images indicates that the proposed Simplified Active Calibration method estimates the parameters of a camera with low error rates that can be used as an initial guess for further non-linear refinement procedures. Simplified Active Calibration can be employed in real-time environments for automatic calibrations given the proposed closed-form solutions. 
\end{abstract}


\end{frontmatter}


\section{Introduction}\label{sec:introduction}
Camera calibration is an essential step in many 3D computer vision applications where we need to calculate how the 3D world is projected onto a 2D image. Camera calibration aims not only to estimate the intrinsic camera parameters such as focal length, center of projection, pixel skew and aspect ratio, but also the camera motions, i.e., the rotation and translation of the camera.

In order to calibrate a camera, conventional calibration methods need to acquire some information from the real 3D world using calibration objects such as grids, wands, LEDs, or even by adding augmented reality markers to a camera \cite{zhao2018marker}. This imposes a major limitation on the calibration task since the camera can be calibrated only in off-line and controlled environments. To address this issue, Maybank and Faugeras \cite{maybank1992theory,faugeras1992camera} proposed the so-called \textit{self-calibration} approach in which they used the information of matched points in several images taken by the same camera from different views instead of using known 3D points (calibration objects). In their two-step method, they first estimated the epipolar transformation from three pairs of views, and then linked it to the image of an absolute conic using the Kruppa equations \cite{maybank1992theory}. Not long after the seminal work of Maybank and Faugeras, Basu proposed the idea of Active Calibration \cite{basu1993active2, basu1993active} in which he included rotations of a camera and eliminated point-to-point correspondences.

An active environment, can change the characteristics of a problem. For instance, an ill-posed and nonlinear problem for a passive observer can become well-defined and linear for an active observer \cite{aloimonos1988active}. Thus, to successfully calibrate the camera, Active Calibration needs to control the camera motion. This makes it a perfect choice in on-line platforms like robotics or surveillance where the internal parameters might change due to focusing, zooming, or mechanical and thermal variations of the environment surrounding a camera. Therefore, knowing the motion of the camera is essential in Active Calibration and as Hartley stated  \cite{hartley1997self}, ``simplifies the calibration task enormously.'' Another advantage of Active Calibration is its closed-form strategies that calculate the intrinsic parameters through only two pairs of images taken after panning and tilting the camera.

Other works \cite{du1993self,dron1993dynamic,stein1995accurate} that used known camera motions have also been published almost at the same time. Since the method of Maybank and Faugeras needed high accuracy in the computations  \cite{hartley1994self,hartley1997self}, complicated rectification processes, and also because of the unavailability of the epipolar structure in the scenes taken from a fixed point, Hartley \cite{hartley1994self} proposed a method for self-calibrating a camera with constant intrinsics using projective distortions of several pure camera rotations. Inspired by Hartley's work, Agapito et al. proposed a self-calibration method for cameras that freely rotate while changing their internal parameters by zooming \cite{agapito2001self}. The notion of varying intrinsics has also been considered in \cite{pollefeys1999self} but no assumption about the camera motion has been made. Research in this area has expanded since the emergence of cell phone cameras capable of measuring the camera motions with Gyroscope and Inertial Measurement Unit (IMU) and Pan-Tilt-Zoom (PTZ) cameras. Given a close approximation of the camera motion, several papers proposed new formulations for calibrating non-rotating stationary cameras \cite{elamsy2014self}, and cameras with known motion \cite{frahm2003camera,frahm2003cameraCal,frahm2003robust}. Some studies calibrate the camera by having specific type of control over camera rotations \cite{knight2003linear,hua2000new,junejo2008practical,wan2010self}. More recent methods proposed self-calibration formulations that include camera lens distortions \cite{wu2013keeping,galego2012auto,sun2016camera}. Also, some researchers expanded the self-calibration formulation to robotic camera networks \cite{heng2015self} or used a camera rotation observed by another camera as the pattern to calibrate the observing camera \cite{bruckner2014intrinsic}. Human motion has also been considered as a way to deduce the camera parameters \cite{tresadern2008camera}.
\begin{figure}[h]\centering
	\includegraphics[width=\linewidth]{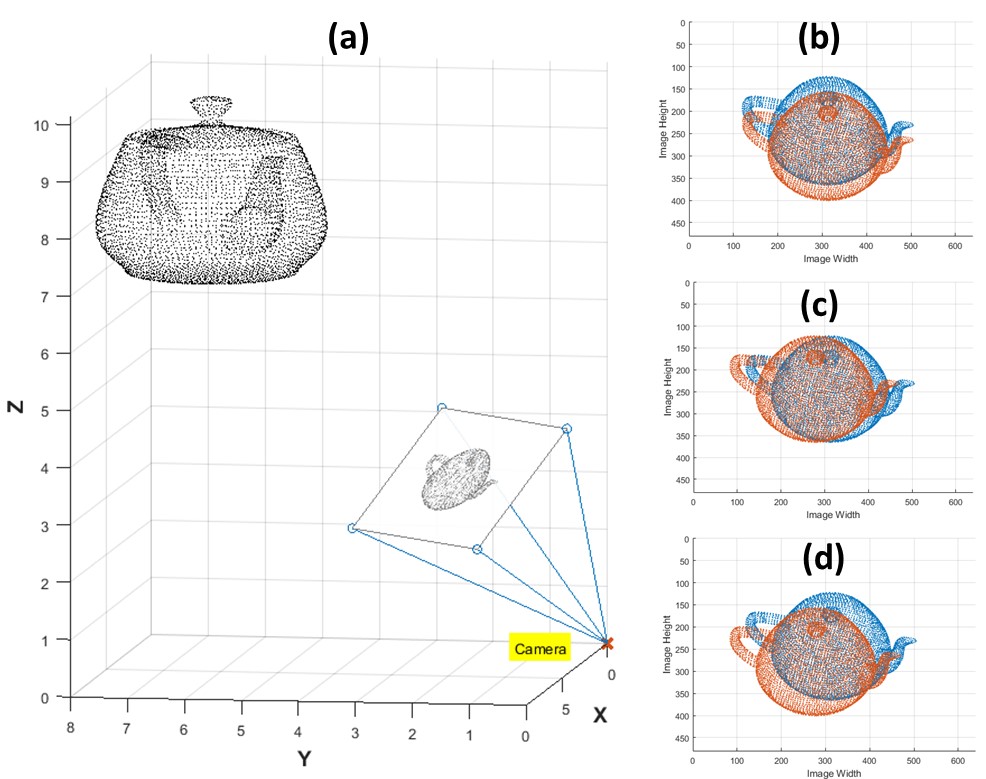}	
	\caption{3D scene and the simulated camera. \textbf{a)} A teapot in the 3D scene and its projected image on the simulated camera. \textbf{b)} The projected image of the teapot on the camera before (blue teapot) and after (red teapot) tilting the camera by $2.5^{\circ}$. \textbf{c)} The projected image of the teapot on the camera before (blue teapot) and after (red teapot) panning the camera by $2.5^{\circ}$. \textbf{d)} The projected image of the teapot on the camera before (blue teapot) and after (red teapot) panning the camera by $2.5^{\circ}$ and then tilting the camera by $2.5^{\circ}$.}
	\label{fig::3Dscene}
\end{figure}

The main downside of the Active Calibration strategies (A and B) in \cite{basu1993active,basu1993active2,basu1995active} is that it calculates the camera intrinsics using a component of the projection equation in which a constraint is imposed by the degenerate rotations. For example, after panning the camera the equation derived from vertical variations observed in the new image plane is unstable. Furthermore, the small angle approximation using $\sin(\theta) = \theta$ and $\cos(\theta) = 1$ decreases the accuracy of the strategies when the angle of rotation is not very small. Also, rolling the camera \cite{basu1997active} is impractical (without having a precise mechanical device) because it creates translational offsets in the camera center. In this paper, we propose a Simplified Active Calibration (SAC) formulation in which the equations are closed-form and linear. To overcome the instability caused by using degenerate rotations in Active Calibration, we calculate focal length in each direction separately \cite{faraji2018simplified}. Then, through a mathematical derivation we remove the corresponding degenerate component from the equation. In addition, we do not use small angle approximation by replacing $\sin(\theta) = \theta$ and $\cos(\theta) = 1$. Hence, in our formulation we only refer to the elements of the rotation matrix. Moreover, the proposed method is more practical because it does not require a roll rotation of the camera; only pan and tilt rotations, which can be easily acquired using PTZ cameras, are sufficient.

The rest of the paper is organized as follows. In Section \ref{sec::SAC} we present our proposed Simplified Active Calibration formulation. Section \ref{sec::results} reports and analyzes the results of the proposed method on simulated and real scenes. Finally, our conclusions are drawn in Section \ref{sec::conclusion}.

\section{Simplified Active Calibration}\label{sec::SAC}
Simplified Active Calibration (SAC) has been inspired by the novel idea of approximating the camera intrinsics using small rotations of the camera which was initially proposed in \cite{basu1993active2, basu1993active} and extended in \cite{basu1995active, basu1997active}. Imposing three constraints on the translation of the camera generates a pure rotation motion. In addition, using small rotation angles provides a condition suitable for ignoring some non-linear terms in order to estimate the remaining linear parameters. The estimated intrinsics can then be used as an initial guess in the non-linear refinement process.

Generally, SAC can be used in any platform in which information about the camera motion is provided by the hardware, such as in robotic applications where the rotation of the camera can be extracted from the inertial sensors or in surveillance control softwares that are able to rotate the PTZ cameras by specific angles. Having access to the rotation of the camera, we propose a 3-step process to calibrate the camera. In the first step, we present a closed-form solution to calculate an approximation of the focal length in the $v$ direction ($f_v$) using an image taken after a pan rotation of the camera, assuming that $v$ and $u$ represent the two major axes of the image plane. In the second step, we estimate the focal length of the camera in the $u$ direction ($f_u$) using an image taken after a tilt rotation of the camera. The third step consists of forming a system of linear equations to estimate the location of the principal point ($v_0,u_0$) in the image. Now, we have estimates for the four main components of the intrinsic matrix, namely $f_v, f_u, v_0$ and $u_0$. Thus, we require three pairs of images, one taken before and after a small pan rotation, one taken before and after a small tilt rotation, and one taken before and after a small pan-tilt rotation. 

\subsection{Rotation Formulation}
Throughout the rest of the paper, we formulate the rotation of the camera by the Euler angles in which every angle in the 3D coordinate system represents the amount of rotation about one of the coordinate axes and is denoted by a separate $3 \times 3$ matrix. The final rotation matrix is thus computed using $\mathbf{R} = \mathbf{R}_z \mathbf{R}_y \mathbf{R}_x$ where $\mathbf{R}_x$, $\mathbf{R}_y $, and $\mathbf{R}_z$ denote the rotations about $x$, $y$, and $z$ respectively. This formulation implies that the resulting matrix $\mathbf{R}$ has three degrees of freedom. Also, the elements of the final rotation matrix are represented as:
\begin{align}
\mathbf{R} = [r_{ij}]_{3 \times 3}
\end{align} 
Where $i$ denotes the row-wise element indices and $j$ represents the column-wise element indices.

In SAC, it is crucial to know the correct direction of the rotation matrix and its handedness since it has to correspond to the acquired images. For example, ``which rotation matrix corresponds to an image acquired after panning the camera to the left.'' Due to the importance of this issue towards having an elegant formulation and obtaining realistic results, we briefly explain every rotation matrix and its direction which is used throughout the paper.

\subsubsection{Roll}
Roll is a rotation about the $z$-axis, used only in Strategies C and D of the original Active Calibration \cite{basu1997active}. However, SAC does not need a rolled image, because rolling the camera and keeping the principal point fixed at the same time is impractical with current cameras. (Imposing a constraint on $v$ or $u$ while rotating about $z$ is very difficult and creates a translational offset \cite{ji2004self}). Therefore, a $3 \times 3$ identity matrix is used to calculate the final rotation matrix. 

\subsubsection{Pan}
Pan rotation of the camera represents a rotation about the $y$-axis and is computed using the following equation. 
\begin{align}
\mathbf{R}_y = 
\begin{bmatrix}
\cos(\theta_p)&0&-\sin(\theta_p)\\
0&1&0\\
\sin(\theta_p)&0&\cos(\theta_p)\\
\end{bmatrix}
\end{align}
The direction implied by this rotation matrix in our predefined camera model is a clockwise orientation if one uses the right-hand rule. Therefore, rotating the camera to the right indicates a positive angle value. On the other hand, for a rotation to the left side the angle should have a negative sign. 

\subsubsection{Tilt}
By tilt rotation we mean a rotation of the camera about the $x$-axis which can be achieved by the following matrix.
\begin{align}
\mathbf{R}_x = 
\begin{bmatrix}
1&0&0\\
0&\cos(\theta_t)&\sin(\theta_t)\\
0&-\sin(\theta_t)&\cos(\theta_t)\\
\end{bmatrix}
\end{align}
Unlike the pan rotation, tilt orientation is counter-clockwise considering the right-hand rule. So, if the camera rotates upward the angle is positive and if it rotates downward the angle of the rotation is negative.

\subsection{Camera Model}
We assume that the camera is located at the origin of the Cartesian coordinate system and is looking at distance $z=f$ where the principal point is specified. It should be noted that $f$ represents the focal length of the camera. Furthermore, the principal axis coincides with the $z$-axis, and the image plane is perpendicular to the principal axis. A point on the normalized camera coordinates is denoted by $\mathbf{x} = [x\;y\;1]^T $. Also, the column ($v$) and row ($u$) coordinate axes of the reference image plane are parallel to the $x$-axis and the $y$-axis of the camera, respectively. The relation between points in the normalized camera coordinates and the image points is as follows:
\begin{align}
&v = m_vx +v_0\\
&u = -m_uy+u_0
\end{align}
Where $m_v$ and $m_u$ represent the width and height of the pixels, respectively and $(v_0,u_0)$ are the location coordinates of the principal point in the image.

Every 3D point $\mathbf{X} = [X\;Y\;Z]^T$ in the world that is visible to the camera can be projected onto a specific point $\mathbf{u} = [v\;u\;1]^T $ of the image plane and can be calculated using the camera intrinsic matrix.
\begin{align}
\mathbf{K} = 
\begin{bmatrix}
f_v & s & v_0\\
0 & -f_u & u_0\\
0 & 0 & 1\\
\end{bmatrix}
\end{align}
Where $f_v = fm_v$ is the focal length of the camera in the $v$ direction (in pixels), $f_u = fm_u$ represents the focal length of the camera in the $u$ direction. With modern cameras it is reasonable to assume that image pixels are square and so the value of the camera skew ($s$) is zero.

Also, any camera transformation is equivalent to a similar transformation of the scene but in the opposite direction  \cite{kanatani1987camera}. For stationary cameras that freely rotate but stay in a fixed location, the camera transformation is only modeled by its rotation. In other words, the translation of the camera is zero. Therefore, every point $\mathbf{u}$ in an image seen by a stationary camera is transformed to a point $\mathbf{u'}$ in another image taken after camera rotation. The mathematical relationship between $\mathbf{u}$ and $\mathbf{u'}$ is thus represented by:
\begin{align}\label{eq::generalPrjEqn}
w\mathbf{u'}= \mathbf{K}\mathbf{R}^{T}\mathbf{K}^{-1}\mathbf{u}
\end{align}
Where $w$ is the scale of the projection and represents the depth of the point. It should be noted that $\mathbf{R}^{T} = \mathbf{R}^{-1}$ because the rotation matrix is orthonormal.
\subsection{Focal Length in the v Direction}\label{sec::fv}
An image taken after a pan rotation of a camera provides a very straightforward formulation to estimate the focal length of the camera. In fact, it imposes two constraints on the camera rotations around the $x$ and $z$ axes. The resulting projection equation after substituting $\mathbf{R_y}$ for $\mathbf{R}$ is:
\begin{align}\label{eq::fvPrj}
w\mathbf{u'}= \mathbf{K}\mathbf{R}_y^{T}\mathbf{K}^{-1}\mathbf{u}
\end{align}
$\mathbf{R}_y^{T}$ has only one DoF which is the angle of rotation around the $y$-axis. After expanding and simplifying Eq.\ref{eq::fvPrj} and eliminating the scale of projection, the following direct projection equations are obtained.
\begin{align}\label{eq::fvPrjv}
v' = \dfrac{r_{11}(v-v_0)+r_{31}f_v}{r_{13}\dfrac{v-v_0}{f_v}+r_{33}} + v_0 
\end{align}
\begin{align}\label{eq::fvPrju}
u' = u_0 - \dfrac{u_0-u}{r_{13}\dfrac{v-v_0}{f_v}+r_{33}} 
\end{align}
Where $r_{ij}$ is an element of $\mathbf{R}^T_y$ at row $i$ and column $j$. After simplification of Eq.\ref{eq::fvPrju}:
\begin{align}\label{eq::fvPrjSimpleU}
\dfrac{v-v_0}{f_v} = \dfrac{\dfrac{u_0-u}{u_0-u'}-r_{33}}{r_{13}}
\end{align}
Note that after a pure pan rotation, the $u$ coordinates of the new image will not be affected by the transformation. (The reader is referred to \cite{junejo2012optimizing} for a detailed explanation and analysis about this fact.) In other words, image pixels only move horizontally and so the rate of change in the $u$ direction before and after the pan rotation is close to one, viz:
\begin{align}\label{eq::fvPrjURate}
\dfrac{u_0-u}{u_0-u'} \approx 1
\end{align}
Substituting Eq.\ref{eq::fvPrjURate} into Eq.\ref{eq::fvPrjSimpleU} and then replacing the equation obtained for the term $\dfrac{v-v_0}{f_v}$ in the Eq.\ref{eq::fvPrjv}, we have:
\begin{align}\label{eq::fvPrjvSubs}
v' \approx \dfrac{r_{11}(v-v_0)+r_{31}f_v}{r_{13}\dfrac{1-r_{33}}{r_{13}}+r_{33}} + v_0 
\end{align}
The above substitution changes the value of the denominator to 1 and hence simplifies the whole projection equation.
\begin{align}\label{eq::fvPrjvFvV0}
v'-r_{11}v \approx r_{31}f_v +(1-r_{11}) v_0
\end{align}
Since Eq.\ref{eq::fvPrjvFvV0} is linear, one might think that it can be solved by constructing a linear system of equations using the matched points from two images taken after the pan rotations of the camera. Unfortunately, the equation is numerically unstable because the value of $1-r_{11}\approx 0$ which causes ambiguity in calculating the shift in the principal point \cite{agapito2001self}. In short, we cannot calculate the location of the principal point in the $v$ direction from a camera rotated purely around the $y$ axis. Knowing that the principal point is close to the center of the image $(c_u=h/2,c_v=w/2)$, where $h$ and $w$ represent the image height and width respectively, we replace $v_0$ with $c_v$ in Eq.\ref{eq::fvPrjvFvV0}. Thus, we can derive a suitable linear equation to estimate the focal length in the $x$ direction from an image taken after a pan rotation.
\begin{align}\label{eq::fvFinal}
f_v \approx \dfrac{v'-r_{11}v- (1-r_{11})c_v}{r_{31}}
\end{align}  
Eq.\ref{eq::fvFinal} needs only one point $v$ in the reference image that corresponds to $v'$ in the transformed image. If there are more point correspondences, we can easily use the average of these points to obtain more robust results.
\subsection{Focal Length in the u Direction}
So far, we could estimate $f_v$ by the information provided from an image taken after a pan rotation. We repeat the same procedure to approximate $f_u$. This time we need an image taken after a pure tilt rotation of the camera. Thus, the projection equation is characterized by replacing $\mathbf{R}$ with $\mathbf{R}_x$ and relating the coordinates of a point in the reference image $\mathbf{u}$ and a point in the tilted image $\mathbf{u}'$ by:  
\begin{align}\label{eq::fyPrjx}
v' = \dfrac{v-v_0}{r_{23}\dfrac{u_0-u}{f_u}+r_{33}} + v_0 
\end{align}
\begin{align}\label{eq::fyPrjy}
u' = u_0 - \dfrac{r_{22}(u_0-u)+r_{32}f_u}{r_{23}\dfrac{u_0-u}{f_u}+r_{33}}
\end{align}
Following the same reasoning as in Section \ref{sec::fv}, a closed-form solution to estimate the focal length of the camera in the $u$ direction is obtained by:
\begin{align}\label{eq::fuFinal}
f_u \approx \dfrac{r_{22}u - u' + (1-r_{22})c_u}{r_{32}}
\end{align}

\subsection{Principal Point}
To estimate the location of the principal point we need to impose one constraint on the rotation matrix which can be achieved by preventing the camera from rotating around the $z$-axis. In real applications the easiest way is to mount the camera on a tripod. In case of working in a robotic environment or with a PTZ camera, controlling roll rotation is straightforward since the camera has already been mounted or fixed. Therefore, we match an image taken after a pan and tilt rotation of the camera with the reference image and use the acquired point correspondences to estimate the location of the principal point.

Following the general projection equation in Eq.\ref{eq::generalPrjEqn}, the direct equations for relating the location of a point in the reference image to its matched point in the transformed image are described by:
\begin{align}\label{eq::generalPrjV}
v' = \dfrac{r_{11}(v-v_0)+r_{21}(u_0-u)\dfrac{f_v}{f_u}+r_{31}f_v}{r_{13}\dfrac{v-v_0}{f_v}+r_{23}\dfrac{u_0-u}{f_u}+r_{33}}+v_0
\end{align}
\begin{align}\label{eq::generalPrjU}
u' = u_0 - \dfrac{r_{12}(v-v_0)\dfrac{f_u}{f_v}+r_{22}(u_0-u)+r_{32}f_u}{r_{13}\dfrac{v-v_0}{f_v}+r_{23}\dfrac{u_0-u}{f_u}+r_{33}}
\end{align}
Where except for $v_0$ and $u_0$, all other terms are known. After simplifying the equations and collecting the coefficients of various powers of $v_0$ and $u_0$, we see that the equations are nonlinear due to the presence of two terms $-r_{13}f_v^{-1}v_0^2-r_{23}f_u^{-1}v_0u_0$ and $-r_{23}f_u^{-1}u_0^2-r_{13}f_v^{-1}v_0u_0$ in Eq.\ref{eq::generalPrjV} and Eq.\ref{eq::generalPrjU} respectively. Nevertheless, the equations can be solved using any nonlinear solver such as Levenberg-Marquardt. But, in order to let the nonlinear solver converge towards the true global minimum, we first need a reasonable initial guess. Here, we propose a method to linearize Eq.\ref{eq::generalPrjV} and Eq.\ref{eq::generalPrjU} to achieve a close estimation of the location of the principal point when the focal lengths and the camera rotations are known.

A feasible approach to linearize the projection equations is to decrease the contributions of the two above-mentioned nonlinear terms in the equations and then eliminate the terms from the equations. Decreasing the value of the nonlinear terms depends on two factors, namely $r_{13}$ and $r_{23}$ which are the elements of the rotation matrix and the value of the unknowns which are $v_0$ and $u_0$. The former coefficients have already been reduced due to our initial assumption of rotating the camera by small angles. On the other hand, we will show that we can reduce $v_0$ and $u_0$ to smaller values by manipulating the scale of these points' coordinates.

Estimating the principal point by a nonlinear algorithm is known to be arduous since it tends to fit to noise \cite{agapito2001self}. Due to this sensitivity to noise, researchers have taken advantage of including some prior knowledge about the distribution of the principal point. It is reasonable to expect that the principal point is close to the center of the image. This prior knowledge is the basis of the \textit{Maximum a Posteriori Estimation} employed in \cite{agapito2001self} to alter the cost function of the minimization problem. In order to arrive at a linear system, we employ the same idea.  Specifically, we assume that the principal point is only slightly shifted from the center of the image.
\begin{align}
&v_0 = c_v + \delta_v \nonumber \\
&u_0 = c_u + \delta_u
\end{align}
Where $\delta_v$ and $\delta_u$ represent the amount of shift in the $v$ and $u$ directions, respectively and $(c_v,c_u)$ are the coordinates of the center of the image. Following this change, we replace each $v-v_0$ term with $\hat{v}-\delta_v$ and every $u_0-u$ with $\hat{u}+\delta_u$ in Eq.\ref{eq::generalPrjV} and Eq.\ref{eq::generalPrjU} where $\hat{v}=v-c_v$ and $\hat{u}=c_u-u$. Therefore, after some simplifications, the general projection equations can be rewritten based on the new variable substitutions.

\begin{align}
&G\delta_v^2 + H\delta_v\delta_u + (A+I-G\hat{v}')\delta_v +(B-H\hat{v}')\delta_u = I\hat{v}'-C\\
&-H\delta_u^2 - G\delta_v\delta_u + (D-G\hat{u}')\delta_v + (E-I-Hu')\delta_u = I\hat{u}'-F
\end{align}
Where,
\begin{align}
&A = -r_{11}    \;\;\;\;\;\; , \;\;\; B=r_{21}\dfrac{f_v}{f_u}  \nonumber \\
&C = r_{11}\hat{v}+r_{21}\hat{u}\dfrac{f_v}{f_u}+r_{31}f_v  \nonumber \\
&D = -r_{12}\dfrac{f_u}{f_v} \;\;\; , \;\;E = r_{22} \nonumber \\
&F = r_{12}\hat{v}\dfrac{f_u}{f_v}+r_{22}\hat{u}+r_{32}f_u \nonumber \\
&G = -\dfrac{r_{13}}{f_v} \;\;\;\;\;\; , \;\;\; H=\dfrac{r_{23}}{f_u} \nonumber \\
&I = r_{13}\dfrac{\hat{v}}{f_v} + r_{23}\dfrac{\hat{u}}{f_u}+r_{33} \nonumber
\end{align}

By including our prior knowledge about the image center into the equations, we significantly reduce the values of nonlinear terms and allow them to be ignored. Once the nonlinear terms are removed, a linear system of equations can be constructed using the detected point correspondences. 
\begin{align} \label{eq::principalPointFinal}
\begin{bmatrix}
A+I_1-Gv_1' &B-Hv_1'\\
\vdots & \vdots \\
A+I_n-Gv_n' &B-Hv_n'\\
D-\hat{u}_1'G & E-I_1-Hu_1'\\
\vdots & \vdots \\
D-\hat{u}_n'G & E-I_n-Hu_n'\\
\end{bmatrix}
\begin{bmatrix}
\delta_v\\
\delta_u
\end{bmatrix} =
\begin{bmatrix}
\hat{v}_1'I_1-C_1\\
\vdots \\
\hat{v}_n'I_n-C_n\\
\hat{u}_1'I_1-F_1\\
\vdots \\
\hat{u}_n'I_n-F_n\\
\end{bmatrix}
\end{align}
Where $(.)_i$ represents using coordinates of the $i$th point in the corresponding term and $n$ is the number of correspondences. As shown above, the system of equations is constructed in the form $A_{[n\times 2]}\mathbf{x}=\mathbf{b}_{[n\times 1]}$ and can be easily solved using a least square method or any linear solver \footnote{MATLAB can solve the equation system by the command $A\backslash \mathbf{b}$.}. However, since there are two unknowns, by detecting only one point correspondence we are able to solve the system for $\delta_x$ and $\delta_y$. For better estimates, one can use more point correspondences.
\begin{figure}[h]\centering
	\begin{tabular}{@{}c@{}c@{}}
		\includegraphics[width=0.5\linewidth]{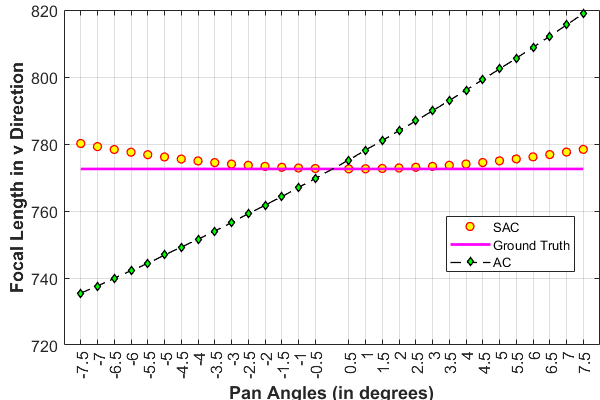} &  \includegraphics[width=0.5\linewidth]{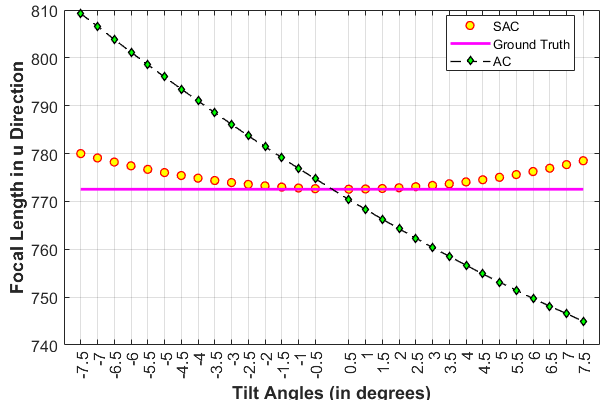}\\
	\end{tabular}	
	\caption{Focal length calculated in the $v$ and $u$ directions using Active Calibration Strategy B (AC) versus SAC for various angles of rotations. In SAC we only use one point correspondence.}
	\label{fig::err_focal}
\end{figure}

\section{Results and Analysis}
\label{sec::results}
In order to better understand the performance of the proposed simplified active calibration, we perform several experiments to clarify how the method works for various rotation angles. We show that rotating by small angle is crucial to obtain good results. 

Based on our proposed method, focal length in the $v$ and $u$ directions can be estimated using Eq.\ref{eq::fvFinal} and Eq.\ref{eq::fuFinal}, respectively. Only one point correspondence is required to calculate the focal length. Fig.\ref{fig::err_focal} shows the focal lengths estimated using various pan and tilt angles. It can be seen that when the pan and tilt angles are small, the estimated focal lengths are very close to the ground truth.

\begin{figure}[h]\centering
	\begin{tabular}{c@{}c}
		\includegraphics[width=0.4\linewidth]{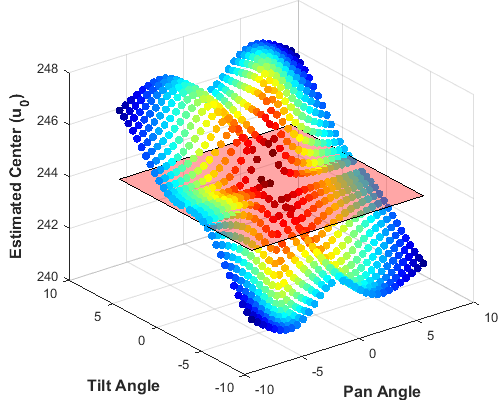} &  \includegraphics[width=0.5\linewidth]{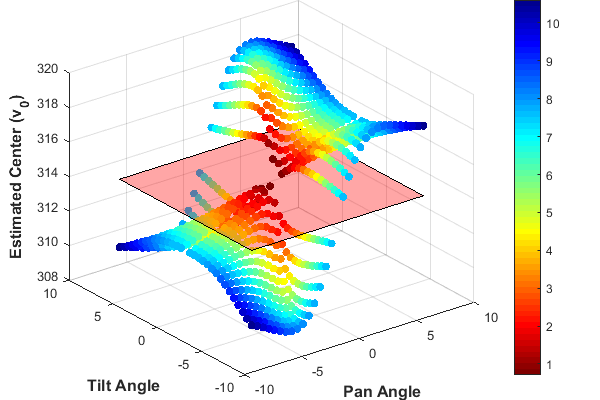}\\
		\textbf{(a)} & \textbf{(b)}\\
		\includegraphics[width=0.45\linewidth]{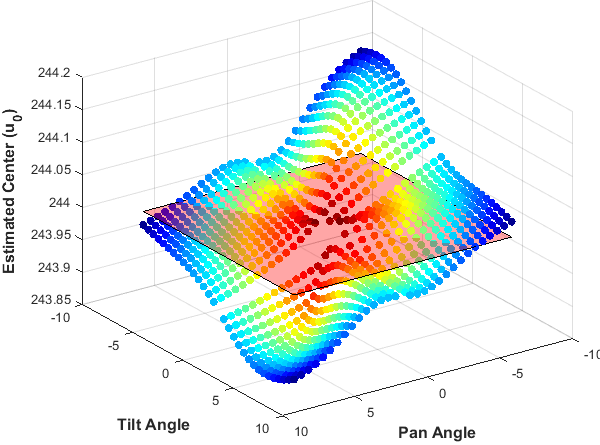} &  \includegraphics[width=0.5\linewidth]{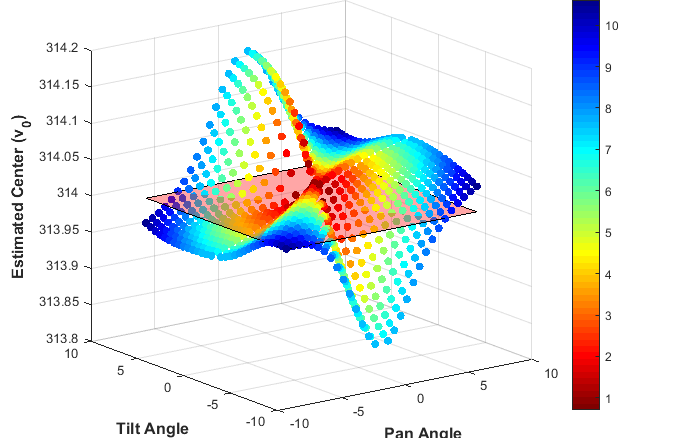}\\
		\textbf{(c)} & \textbf{(d)}\\
		
	\end{tabular}
	\caption{Coordinates of the principal points calculated after various pan/tilt rotations of random 3D points. Colors are distributed based on the $L^2$ norm of the pan and tilt angles. \textbf{a)} Shows the values obtained for $u_0$ when inaccurate focal lengths ($f_u =  774.71$ and $f_v=771.18$) are used. $MSE(u_0)=1.49$ pixels for all combinations of pan and tilt angles. \textbf{b)} Shows the values obtained for $v_0$ when inaccurate focal lengths ($f_u =  774.71$ and $f_v=771.18$) are used. $MSE(v_0)=2.30$ pixels for all combinations of pan and tilt angles. \textbf{c)} Shows the values obtained for $u_0$ when accurate focal lengths ($F_u = F_v = 772.55$) are used. $MSE(u_0)=0.05$ pixels for all combinations of pan and tilt angles. \textbf{d)} Shows the values obtained for $v_0$ when accurate focal lengths ($F_u = F_v = 772.55$) are used. $MSE(v_0)=0.04$ pixels for all combinations of pan and tilt angles. The red plane specifies the ground truth.}
	\label{fig::err_principalPoint}
\end{figure}

The magnitude of the pan and tilt angles affect estimating the principal point location as well. In fact, two successive rotations (pan and tilt) are required to calculate the center of projection. Thus, in another experiment, we rotate the camera by 900 combinations of pan and tilt rotations (from $-7.5^{\circ}$ to $7.5^{\circ}$), and then use the projected points on the image plane to estimate the principal point location by calculating the proposed formulation (Eq.\ref{eq::principalPointFinal}). Note that we use either the estimated value of $f_v$ and $f_u$ that were calculated in the previous step or the actual focal length. The results are shown in Fig.\ref{fig::err_principalPoint}. We can see that for small rotations, the results obtained by our formulation are close to the real location which has been identified in the figure by a red plane. Even when the focal length is not accurate, a good estimate of the principal point can be found if the pan and tilt rotations are small.

In addition, Fig.\ref{fig::err_principalPoint} shows that the error caused by the angle variation is more negligible than the error caused by inaccurate values of the focal length. Specifically, the Mean Square Error of $u_0$ ($MSE(u_0)$) is 1.49 pixels when inaccurate focal lengths are used (Fig.\ref{fig::err_principalPoint}(a)). By contrast, when the actual focal length is used in Eq.\ref{eq::principalPointFinal}, $MSE(u_0)$ is decreased to 0.05 pixels (Fig.\ref{fig::err_principalPoint}(c)). This reveals the significance of having accurate focal length in calculating the principal point location. A similar analysis is valid for the other axis ($v$) which is shown in Fig.\ref{fig::err_principalPoint}(b) and Fig.\ref{fig::err_principalPoint}(d).

\begin{figure}[h]\centering
	\begin{tabular}{@{}c@{}c}
		\includegraphics[width=0.5\linewidth]{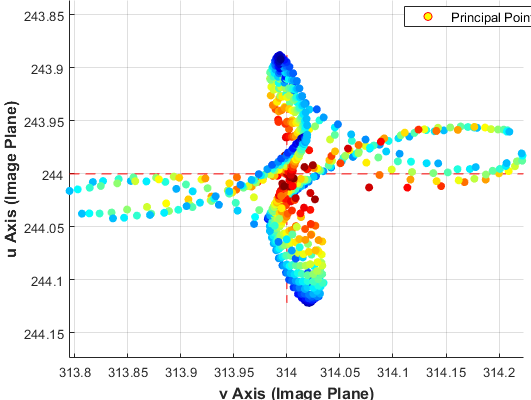} &  \includegraphics[width=0.48\linewidth]{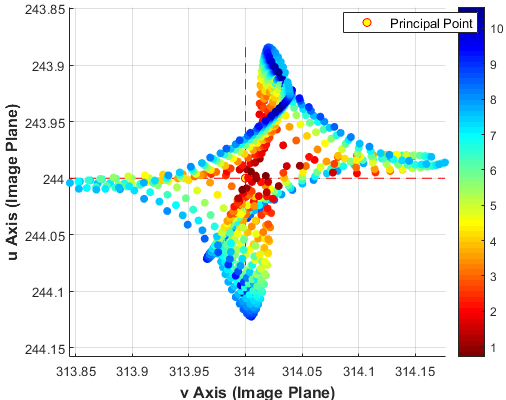}\\
		\textbf{(a)} & \textbf{(b)}\\
	\end{tabular}
	\caption{The estimated locations of the principal point on the image plane for combinations of various rotation angles (from $-7.5^{\circ}$ to $7.5^{\circ}$) using Eq.\ref{eq::principalPointFinal}. Colors are distributed based on the $L^2$ norm of the pan and tilt angles. \textbf{a)} Results for solving with only four point correspondences of the teapot point cloud. \textbf{b)} Results for solving with 500 point correspondences of the random 3D points. The actual principal point location is (314,244).}
	\label{fig::err_vu}
\end{figure}

Knowing how many point correspondences are required to calculate the principal point location (by Eq.\ref{eq::principalPointFinal}) is crucial. Based on our experiments, with only four points that are uniformly distributed in the image, a good estimate of the principal point location can be obtained.  Fig.\ref{fig::err_vu}(a) illustrates principal point locations (on the image plane) obtained by solving Eq.\ref{eq::principalPointFinal} with inaccurate focal lengths and only four point correspondences of the teapot point cloud. Fig.\ref{fig::err_vu}(b) shows the estimated principal points on the image obtained by solving Eq.\ref{eq::principalPointFinal} with inaccurate focal lengths and 500 point correspondences of random 3D points. Both experiments are carried out on 900 combinations of pan and tilt angles which range from $-7.5^{\circ}$ to $7.5^{\circ}$. Pan and tilt angles are included in Fig.\ref{fig::err_vu} by calculating the $L^2$ norm of the angles ($\sqrt{\theta_p^2 + \theta_t^2}$) and assigning meaningful colors to them that range from red (closer to zero) to blue (bigger angles). As can be seen in Fig.\ref{fig::err_vu}(a), when the pan and tilt rotations are small, even with four point correspondences a principal point that is very close to the actual principal point (specified by a red cross) can be calculated. Using more point correspondences we obtain almost similar error distribution, which is shown in Fig.\ref{fig::err_vu}(b).

In another experiment, we calculate the proposed simplified active calibration formulation on 1000 different runs of 500 randomly generated 3D points for small pan ($\theta_p = -0.5^{\circ}$) and tilt ($\theta_t = 0.5^{\circ}$) angles. The order of calculating the intrinsics was specified earlier. The mean and standard deviation of the results obtained are shown in Table \ref{tbl::random3D}. As we can see, our proposed active calibration formulation attains results very close to the ground truth. Specifically, the error in the principal point location is less than one pixel and the error in focal length estimates is less than 2 pixels.
\begin{table}[h]\scriptsize
	\begin{center}
		\caption{Results of the proposed simplified active calibration on 1000 separate 3D random points for various small pan and tilt angles. In the table, GT denotes the Ground Truth and SD represents the Standard Deviation. The error values are in pixels. }
		\label{tbl::random3D}
		
		\begin{tabular}{|@{}c@{}c@{}c|cccc@{}|}
			\hline
			Pan & Tilt &   &	$f_v$  &$f_u$   &$v_0$   &$u_0$ \\
			\hline\hline
			&&GT         &	772.55 &772.55  &314     &244 \\
			\hline
			\multirow{3}{*}{$-0.5^{\circ}$} & \multirow{3}{*}{$0.5^{\circ}$} &		Mean                 &	772.68 &772.57  &314.005  &244.02 \\
			& & SD   &	0.07   &0.01    &0.01     &0.01 \\
			& & Error   &	0.13    &0.02  &0.005    &0.02 \\	
			\hline	
			\multirow{3}{*}{$-0.5^{\circ}$} & \multirow{3}{*}{$1^{\circ}$} &		Mean                 &	772.68 &772.62  &314.03  &244.06 \\
			& & SD   &	0.07   &0.04    &0.07     &0.06 \\
			& & Error   &	0.13    &0.07  &0.03    &0.06 \\	
			\hline	
			\multirow{3}{*}{$1^{\circ}$} & \multirow{3}{*}{$-1^{\circ}$} &		Mean                 &	772.61 &772.76  &314.23  &243.61 \\
			& & SD   &	0.02   &0.09    &0.12     &0.15 \\
			& & Error   &	0.06    &0.21  &0.23    &0.38 \\	
			\hline	
			\multirow{3}{*}{$-1.5^{\circ}$} & \multirow{3}{*}{$1.5^{\circ}$} &		Mean                 &	773.02 &772.73  &314.21  &244.44 \\
			& & SD   &	0.13   &0.07    &0.17     &0.17 \\
			& & Error   &	0.47    &0.19  &0.21    &0.44 \\	
			\hline	
		\end{tabular}
	\end{center}
\end{table}
All things considered, we assessed the proposed Simplified Active Calibration formulation on simulated scenes in ideal situations, i.e., when the 3D rays are not altered due to camera lens distortions and when there is no noise in the scene. We showed that the proposed formulation can estimate the camera intrinsics when the camera rotation is small and pure. In fact, for calculating the focal length we used the so-called ``degenerate camera configuration.'' Moreover, we demonstrated that using small rotations one can compensate for the error caused by inaccuracy in the estimated focal length to find the principal point. In other words, rotating the camera by small angles lessens the influence of inaccurate focal length in calculating the principal point location.
\begin{center}
	\begin{figure*}[h]\centering
		\begin{tabular}{cc}
			\includegraphics[width=0.4\linewidth]{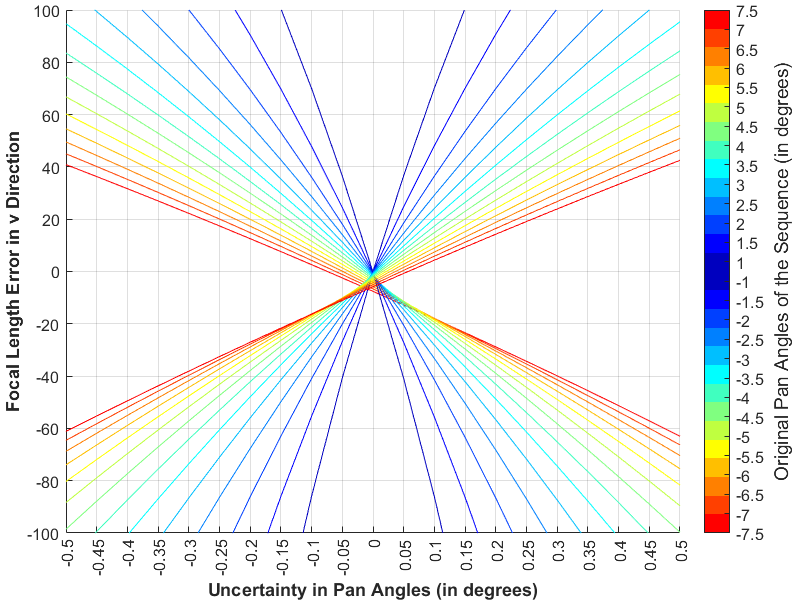} &  \includegraphics[width=0.4\linewidth]{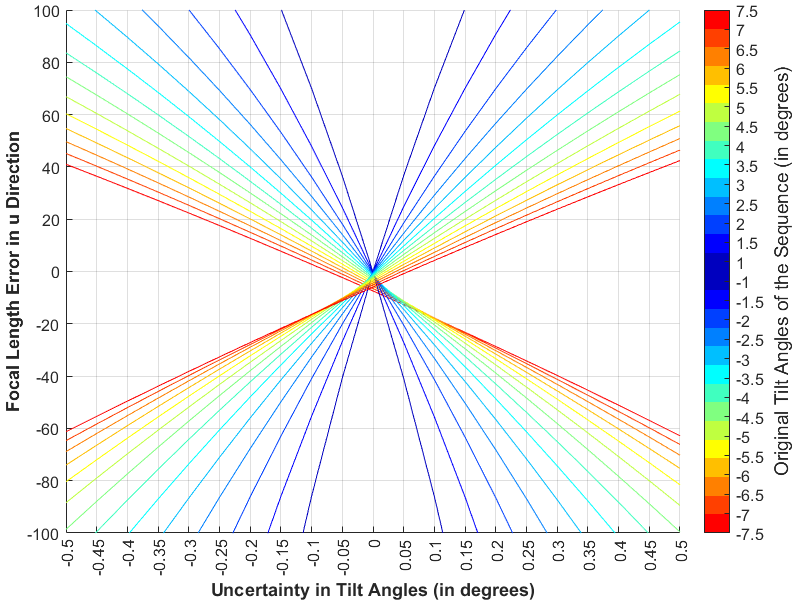}\\
			\textbf{(a)} & \textbf{(b)} \\
			\includegraphics[width=0.4\linewidth]{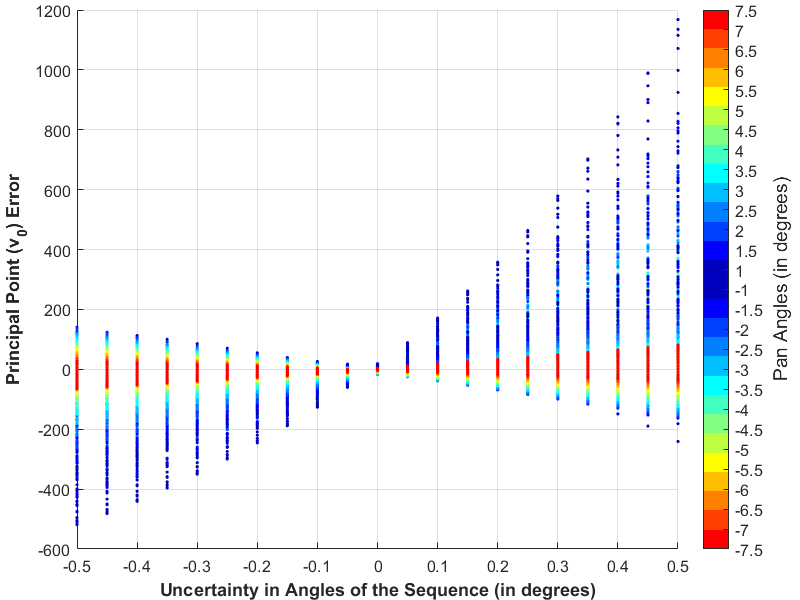} &  \includegraphics[width=0.4\linewidth]{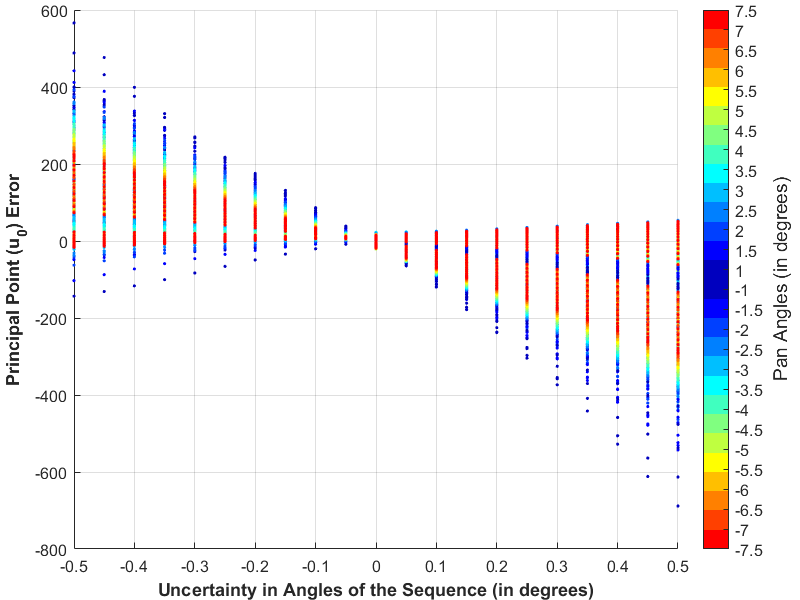}\\
			\textbf{(c)} & \textbf{(d)}\\
		\end{tabular}
		\caption{The error caused by uncertainty in determining the angle of the camera. \textbf{a)} The effects of the uncertainty of the camera pan rotation on calculating the focal length in the $v$ direction by SAC. \textbf{b)} The effects of the uncertainty of the camera tilt rotation on calculating the focal length in the $u$ direction by SAC. \textbf{c)} The effects of the uncertainty of the camera pan and tilt rotation on calculating the $v$ coordinate of the principal points by SAC. \textbf{d)} The effects of the uncertainty of the camera pan and tilt rotation on calculating the $u$ coordinate of the principal points by SAC.}
		\label{fig::err_Ang}
	\end{figure*}		
\end{center}

\subsection{Noise Analysis} 
All of the above-mentioned experiments were done in ideal situations where the angles acquired from the camera and the location of matched points were assumed to be exact. In real-world conditions, however, angles and point correspondences are noisy. In the following sections we try to understand how the proposed method works in real-world conditions where parameters are contaminated by various types of noise. 

\subsection{Angular Uncertainty}
Acquiring the rotation angles requires either specific devices such as gyroscopes or a specially designed camera called a PTZ camera. Even using these devices does not guarantee that the extracted rotation angles are noise-free. To simulate the noisy conditions of a real-world application, we contaminated the angles of the above-mentioned teapot sequences with increasing angular errors. 

While the point correspondences are kept fixed for all of the pan and tilt rotations, we calculate the focal length (Eq.\ref{eq::fvFinal} and Eq.\ref{eq::fuFinal}) and principal point coordinates (Eq.\ref{eq::principalPointFinal}) using contaminated pan and tilt angles. The results are shown in Fig.\ref{fig::err_Ang}. Specifically, Fig.\ref{fig::err_Ang}(a) and Fig.\ref{fig::err_Ang}(b) show the error of our proposed formula for estimating the focal length when the pan and tilt angles are not accurate. Every sequence has been coloured based on its rotation angle, ranging from blue indicating smaller angles to red for larger angles. For focal length estimation, Fig.\ref{fig::err_Ang}(a) and Fig.\ref{fig::err_Ang}(b) illustrate that the sequences taken with smaller angles have steeper slope than the sequences acquired with larger rotation angles. This shows that focal lengths are more sensitive to angular noise when the camera is rotated by smaller angles rather than larger angles.
\begin{center}
	\begin{figure*}[h]\centering
		\begin{tabular}{cc}
			\includegraphics[width=0.4\linewidth]{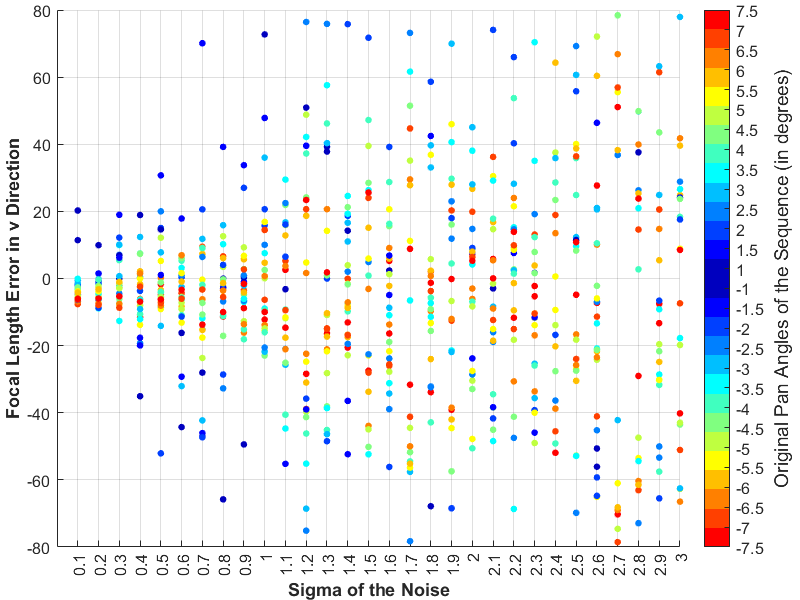} &  \includegraphics[width=0.4\linewidth]{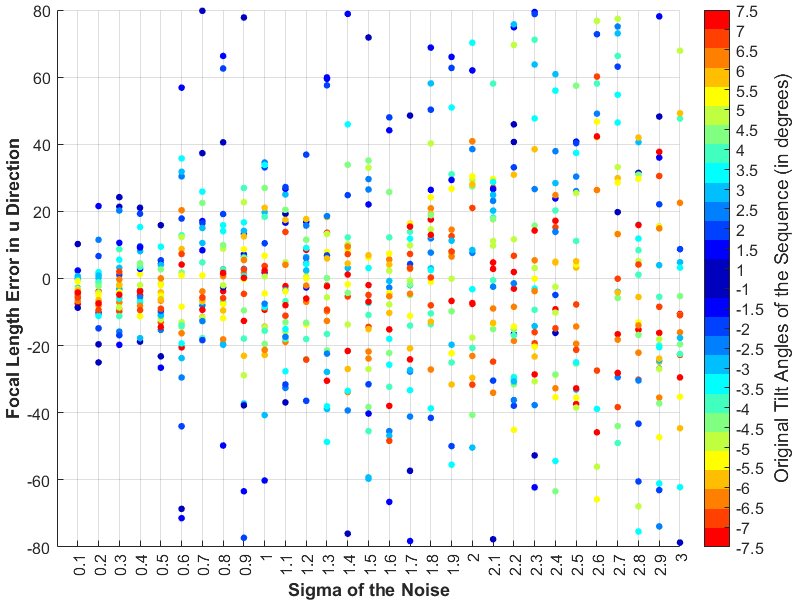}\\
			\textbf{(a)} & \textbf{(b)}\\
			\includegraphics[width=0.4\linewidth]{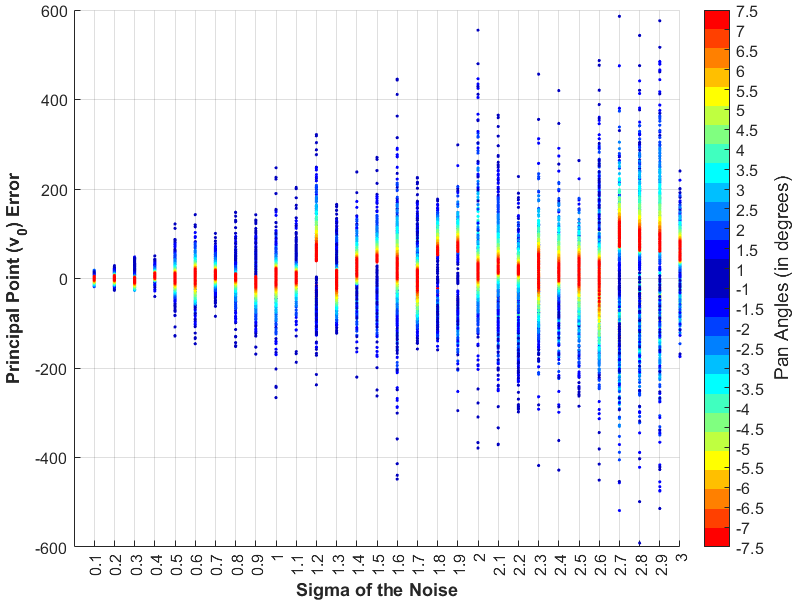} &  \includegraphics[width=0.4\linewidth]{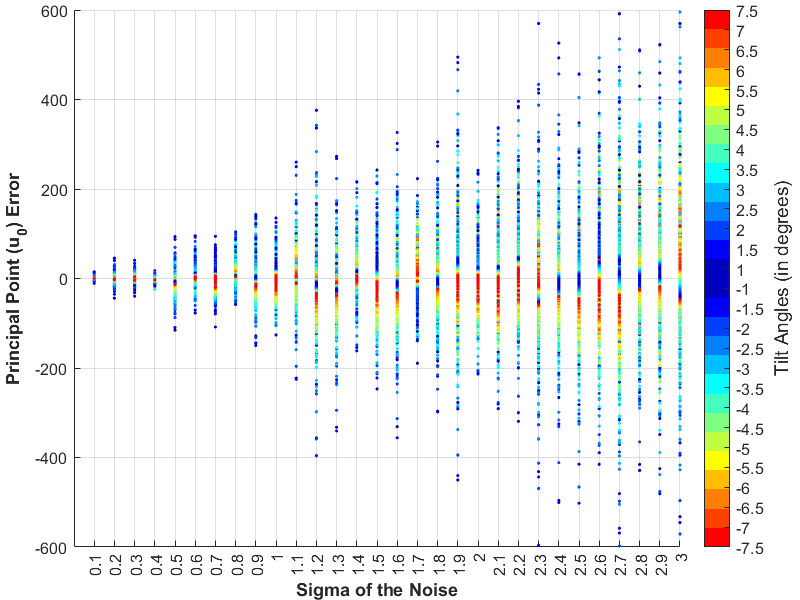}\\
			\textbf{(c)} & \textbf{(d)}\\
		\end{tabular}
		\caption{The error caused by uncertainty in location of points. \textbf{a)} Error of the estimated focal length in the $v$ direction using SAC when the location of the teapot points are disturbed by different values of $\sigma_{pixel}$. \textbf{b)} Error of the estimated focal length in the $u$ direction using SAC under the same conditions as in (a).  \textbf{c)} Error of the estimated $v_0$ using SAC under the same conditions as in (a). \textbf{d)} Error of the estimated $u_0$ using SAC under the same conditions as in (a).}
		\label{fig::err_PC}
	\end{figure*}		
\end{center}

Fig.\ref{fig::err_Ang}(c) and Fig.\ref{fig::err_Ang}(d) demonstrate how uncertainty in angular values affects estimating the principal point coordinates. Similar to the effect of noise on focal lengths, the distribution of the red colours (greater angles) around the zero line in Fig.\ref{fig::err_Ang}(c) and Fig.\ref{fig::err_Ang}(d) indicates that the estimates of the principal point coordinates are less affected by the angular noise when the angle of rotation is not very small. 

Overall, when the camera is rotated by small angles, the influence of the angular noise on SAC equations is significant. On the other hand, SAC tends to use the benefit of rotating the camera by small angles. Therefore, to avoid magnifying the effect of noise it is important not to rotate the camera by very small angles. If the pan and tilt angle of the camera is not very small (usually $< 2^\circ$), the difference in estimated focal lengths will be less than $50$ pixels which are still considered as close initial guesses for further non-linear refinements. Nonetheless, our experiments with real images in Section~\ref{sec::real} reveal that SAC can be used in real situations and the angular noise makes the estimation slightly inaccurate.

\begin{figure*}[h]\centering
	\includegraphics[width=0.5\linewidth]{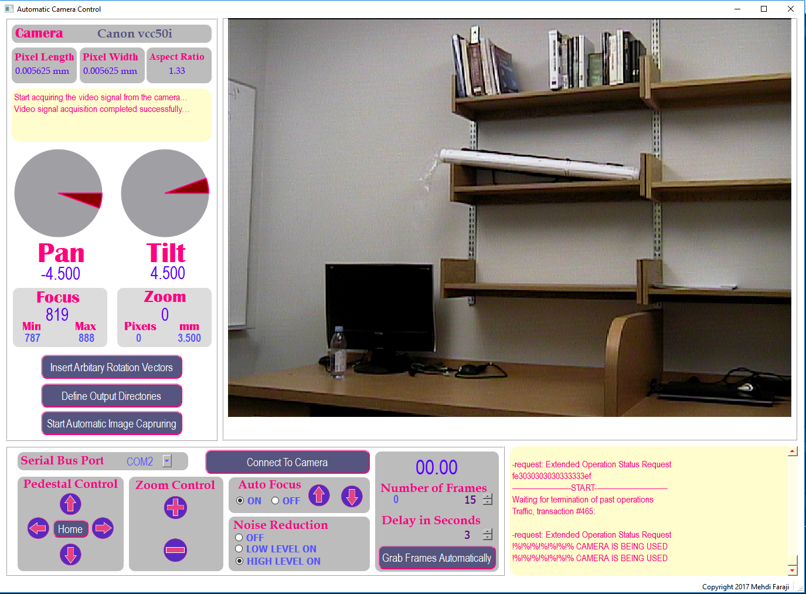} 
	\caption{A screenshot of the designed Automatic Camera Control application that is able to rotate the camera by specific angles about $Y$-axis (pan) and $X$-axis (tilt). The camera is Canon vc-c50i.}
	\label{fig::app}
\end{figure*}

\begin{figure*}[h]\centering
	\begin{tabular}{cccc}
		\includegraphics[width=0.23\linewidth]{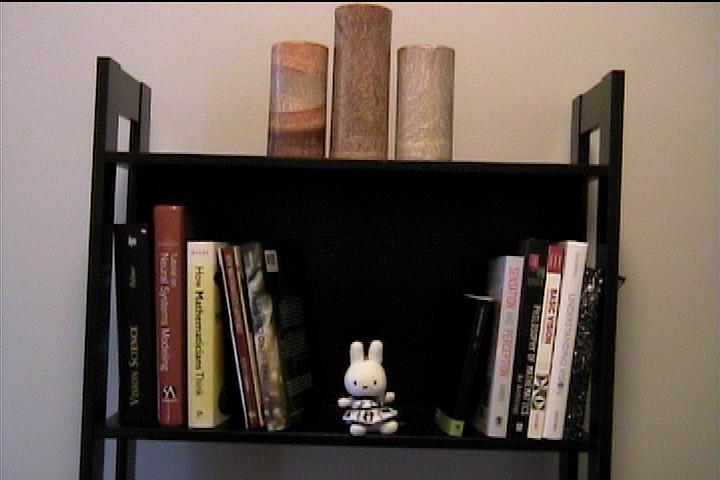} &  \includegraphics[width=0.23\linewidth]{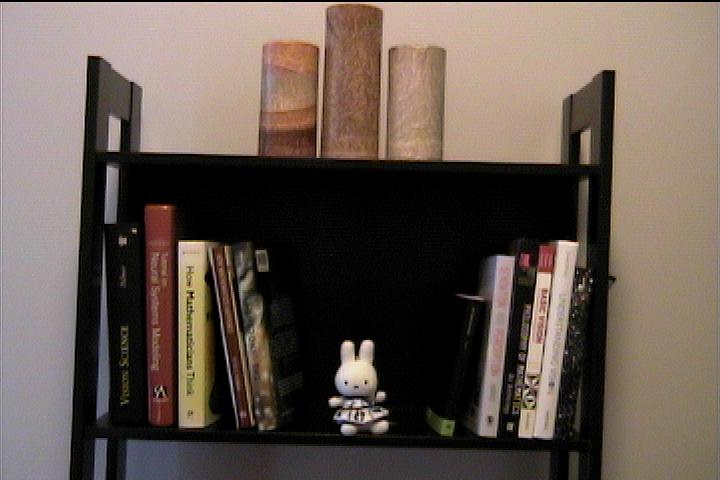}&
		\includegraphics[width=0.23\linewidth]{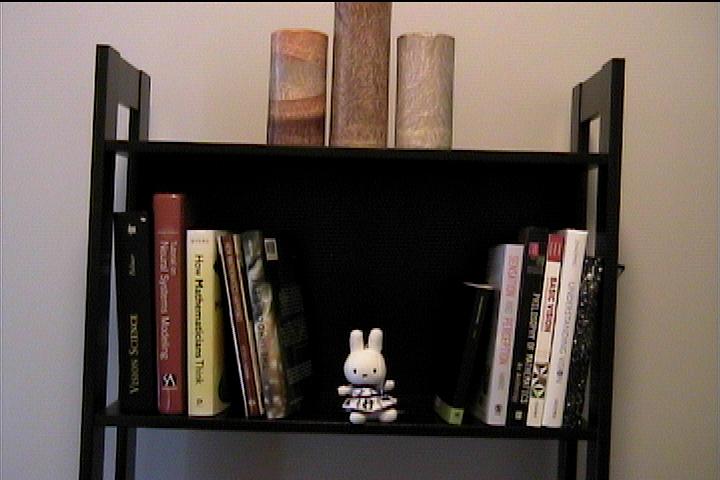} &  \includegraphics[width=0.23\linewidth]{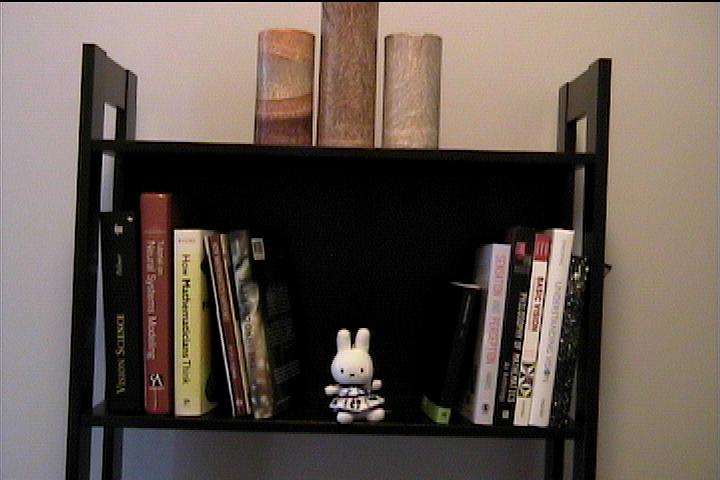}\\
		\textbf{(a)} & \textbf{(b)} &\textbf{(c)} & \textbf{(d)}\\
		\includegraphics[width=0.23\linewidth]{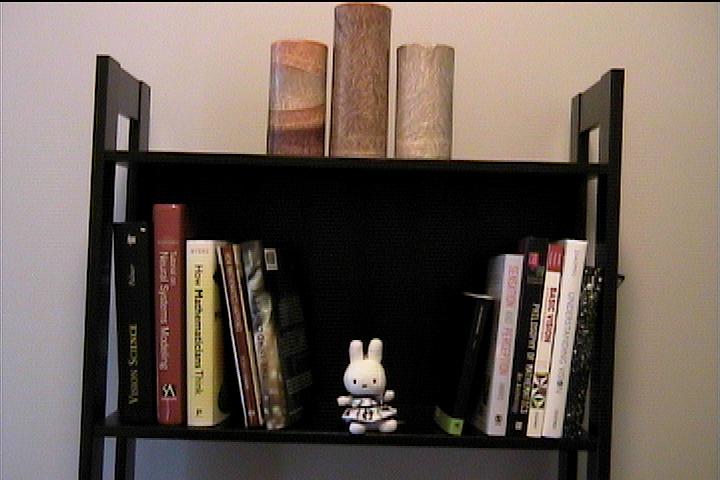} &  \includegraphics[width=0.23\linewidth]{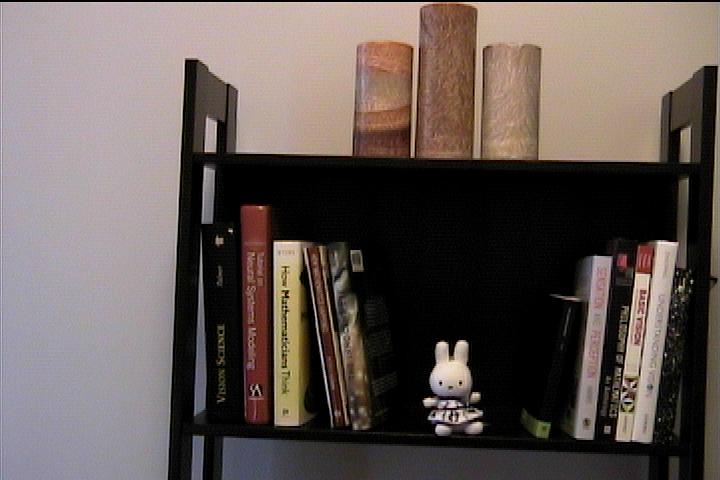}&
		\includegraphics[width=0.23\linewidth]{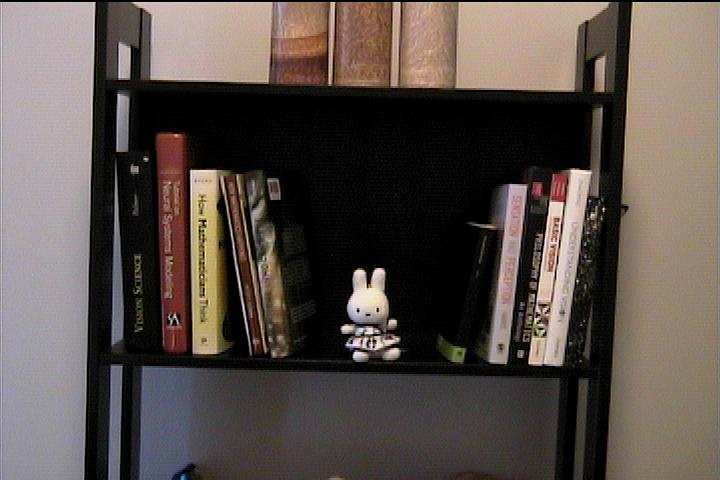} &  \includegraphics[width=0.23\linewidth]{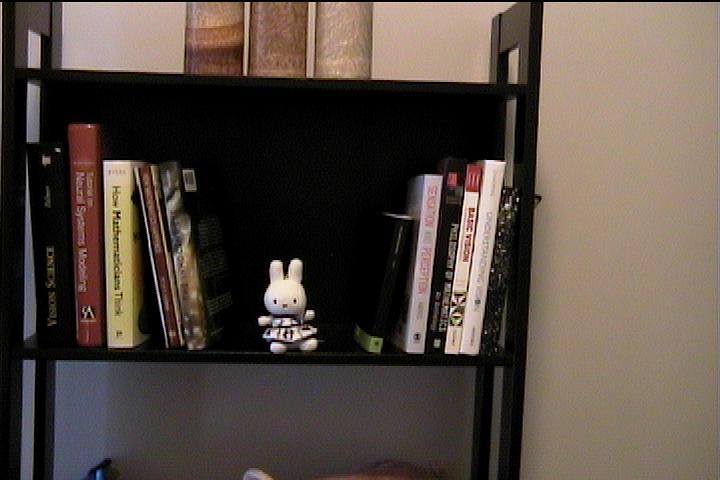}\\
		\textbf{(e)} & \textbf{(f)} &\textbf{(g)} & \textbf{(h)}\\
	\end{tabular}
	\caption{Two sequences of real images used for SAC. Every row represents one sequence. \textbf{a)} A reference image. \textbf{b)} Image taken after panning the camera by $0.5625^\circ$. \textbf{c)} Image taken after tilting the camera by $-0.675^\circ$. \textbf{d)} Image taken after first panning the camera by $0.7875^\circ$ and then tilting the camera by $-0.675^\circ$. \textbf{e)} A reference image. \textbf{f)} Image taken after panning the camera by $-4.6125^\circ$. \textbf{g)} Image taken after tilting the camera by $-4.1625^\circ$. \textbf{h)} Image taken after first panning the camera by $4.3875^\circ$ and then tilting the camera by $-4.6125^\circ$.}
	\label{fig::real_images}
\end{figure*}

\subsection{Point Correspondence Noise}
Another type of noise that affects the SAC equations is the noise in the location of features used for matching. To simulate such conditions, we assume that the location of every teapot point is disturbed by a Gaussian noise with zero mean and variance $\sigma_{pixel}$. Then, we calibrate the camera using SAC for all $\sigma_{pixel}$ in the range of $0$ to $3$. The intrinsic parameters obtained are illustrated in Fig.\ref{fig::err_PC}.

Fig.\ref{fig::err_PC}(a) and Fig.\ref{fig::err_PC}(b) illustrate the influence of pixel noise on the estimation of focal length (Eq.\ref{eq::fvFinal} and Eq.\ref{eq::fuFinal}). Also, Fig.\ref{fig::err_PC}(c) and Fig.\ref{fig::err_PC}(d) show how SAC estimates the coordinate of the principal point (Eq.\ref{eq::principalPointFinal}) in noisy conditions. Colours are distributed based on the rotation angle of the camera and, hence, the distribution of the colours reveals how noise affects the SAC equations. In fact, the high concentration of red, yellow, and orange points around the zero error line in Fig.\ref{fig::err_PC}(a) to (d) reveals that when the angle of the camera rotation is not very small, SAC achieves low-error estimates for focal lengths. This corroborates the claim that very small camera rotations can cause results from the SAC formulations to have high error.

\subsection{Real Images}\label{sec::real}
We studied the proposed SAC formulations on real images as well. We used a Canon VC-C50i PTZ camera that is able to freely rotate around the $Y$-axis (pan) and the $X$-axis (tilt). The camera can be controlled by an application called Automatic Camera Control (created by the first author) that uses a standard RS-232 serial communication to control the camera. Therefore, the required pan and tilt rotation angles can be set in a specific packet and then be written into the camera serial buffer to cause the camera to rotate based on the assigned rotation angles. A screenshot of the ACC application can be seen in Fig.\ref{fig::app}.

Using the above-mentioned procedure, we took 8 sequences of images for evaluating the proposed SAC formulations. Fig.\ref{fig::real_images} shows two sequences of our bookshelf scene. All sequences were taken using a fixed zoom. While keeping the zoom of the camera unchanged, another 30 images were acquired from various viewpoints of a checkerboard pattern. The ground truth for the intrinsic parameters were calculated by applying the method of Zhang \cite{zhang1999flexible} on the checkerboard images.

The performance of SAC formulations on the 8 sequences of real images is reported in Table \ref{tbl::RealImages}. For every sequence, we only used the images in the sequence. For example, to calculate the focal length in the $v$ direction of Sequence 1, we found the point correspondence using \cite{faraji2015erel,faraji2015extremal} between the reference image (Fig.\ref{fig::real_images}(a)) and the image taken after the pan rotation of the camera (Fig.\ref{fig::real_images}(b)). Then, we used only one of the matched points that is closer to the center of the image. Although, we did not include the lens distortion parameter into the SAC formulation (because it creates non-linear equations), we decrease the inaccuracy of the focal length estimates by using a matched point that is closer to the center of the image. Thus, the results are less affected by the lens distortion. A similar procedure was adopted with the image taken after a tilt rotation of the camera (Fig.\ref{fig::real_images}(c)) for calculating the focal length in the $u$ direction of Sequence 1. The location of the principal point was estimated by corresponding the reference image (Fig.\ref{fig::real_images}(a)) to the image taken after panning and tilting the camera (Fig.\ref{fig::real_images}(d)). To solve the linear system of equations of SAC (Eq.\ref{eq::principalPointFinal}), we used all of the matched points.

The errors reported by applying SAC on 8 different sequences of real images in Table \ref{tbl::RealImages} show that inspite of the presence of various types of noise, such as angular uncertainties, point correspondence noise and lens distortion, focal lengths estimated by SAC are close to the results of the method of Zhang \cite{zhang1999flexible}, except when the angles of rotations are very small ($< 1^\circ $). As we discussed earlier, the SAC formulation for estimating the coordinates of the principal point is sensitive to angular noise. Therefore, we can see that sometimes the error in the principal point estimate is increased; for example, in Sequences 3 and 4 of Table \ref{tbl::RealImages}. One can decrease this error by including more matched points (in Eq.\ref{eq::principalPointFinal}) taken after panning and tilting the camera by various angles. 
\begin{center}
	\begin{table*}[ht]\scriptsize
		\caption{Results of the proposed Simplified Active Calibration on 8 sequences of real images. All angles are in degrees. The ``Pan'' column indicates the pan angle of the camera for the first image of the sequence. The ``Tilt'' column represents the tilt angle of the camera for the second image of the sequence. The ``PT Pan'' and ``PT Tilt'' columns denote the pan and tilt angles of the camera for the third image of the sequence, respectively. $\delta_{f_v}$,$\delta_{f_u}$,$\delta_{v_0}$, and $\delta_{u_0}$ are the percentage errors from the corresponding ground truth acquired from calibrating the camera using the method by Zhang \cite{zhang1999flexible}. }
		\label{tbl::RealImages}	
		\begin{tabular}{|ccccccccccccc|}
			\hline
			\#	&	  Pan   &    Tilt    &   PT Pan  &  PT Tilt  &  $f_v$   &  $f_u$   &  $v_0$    & $u_0$    &   $\delta_{f_v}$  &   $\delta_{f_u}$  & $\delta_{v_0}$ & $\delta_{u_0}$ \\
			\hline
			1 &		 0.5625${^\circ}$ &    -0.675${^\circ}$  &   0.7875${^\circ}$  &   -0.675${^\circ}$  &  880.42  &  -999.07 &   299.63  &  276.25  &  15.3  &  5.33  &  13.0  &  1.18  \\
			2 &		   -1.8${^\circ}$ &     2.025${^\circ}$  &   -1.575${^\circ}$  &      1.8${^\circ}$  &  1052.1  &  -966.35 &   394.12  &  215.69  &  1.18  &  1.88  &  5.45  &  10.2  \\
			3 &		 3.0375${^\circ}$ &      -3.6${^\circ}$  &   2.8125${^\circ}$  &  -3.2625${^\circ}$  &  1092.2  &  -983.84 &   380.76  &  206.29  &  5.04  &  3.72  &  11.1  &  21.6  \\
			4 &		-4.6125${^\circ}$ &   -4.1625${^\circ}$  &   4.3875${^\circ}$  &  -4.6125${^\circ}$  &  1067.9  &     -970 &   377.45  &  184.75  &  2.70  &  2.26  &  12.1  &  30.4  \\
			5 &		-7.0875${^\circ}$ &     6.975${^\circ}$  &   -7.425${^\circ}$  &   7.0875${^\circ}$  &  1089.5  &  -989.53 &   355.98  &  254.49  &  4.79  &  4.32  &  7.78  &  5.84  \\
			6 &		-7.9875${^\circ}$ &    -7.425${^\circ}$  &     1.35${^\circ}$  &  -0.7875${^\circ}$  &  1069.6  &  -986.58 &   346.62  &  248.81  &  2.87  &  4.01  &  1.25  &  4.94  \\
			7 &		      9${^\circ}$ &       9.9${^\circ}$  &  -1.0125${^\circ}$  &   1.0125${^\circ}$  &  1074.2  &  -996.25 &   327.39  &  301.77  &  3.31  &  5.03  &  0.31  &  6.04  \\
			8 &		 -6.975${^\circ}$ &   -5.9625${^\circ}$  &   4.3875${^\circ}$  &     -0.9${^\circ}$  &  1064.2  &  -991.71 &      346  &  264.68  &  2.35  &  4.55  &  0.61  &  14.9  \\
			\hline
		\end{tabular}
	\end{table*}
\end{center}

\section{Conclusion}\label{sec::conclusion}
In this paper we presented a new Simplified Active Calibration formulation. Our derivations provided closed-form and linear equations to estimate the parameters of the camera using three image pairs taken before and after panning, tilting, and panning-tilting the camera. 

A basic assumption about the rotation of a fixed camera was made; i.e., to solve the proposed equations, knowing the rotation angles of the camera is necessary. The proposed formulation can be used in practical applications such as surveillance, because in PTZ and mobile phone cameras accessing the camera motion information is straightforward.      

The proposed closed-form formulations for estimating the focal lengths can be solved with only one point correspondence. Finding the correspondence point is straightforward. Following recent developments in feature extractors, one can extract repeatable regions from a pair of images. This is especially useful for applications that favour no point correspondences; where instead of the reference and transferred points in Eq.\ref{eq::fuFinal} and Eq.\ref{eq::fvFinal}, the average of the edge points or the centroid of the region can be used. 

The results of solving our proposed formulations on randomly simulated 3D scenes indicated a very low error rate in estimating the focal lengths and the principal point location. We evaluated our proposed SAC formulation for two different noise conditions, namely angular and pixel noise. The simulated results showed that if the angle of rotation is not very small, the error caused by using SAC formulation is low and can be alleviated by a further non-linear refinement. This conclusion was later verified in our experiment with real images. Our future work will focus on including non-linear parameters into the Simplified Active Calibration equations and use the result of the current study as a close initial guess for an optimization procedure. 

\section{Acknowledgements}
The authors acknowledge the financial support of NSERC, Canada, in making this work possible.



\bibliography{mybibfile}

\end{document}